\documentclass[sigconf]{acmart}

\renewcommand\footnotetextcopyrightpermission[1]{} 
\usepackage{booktabs} 

\setcopyright{none}

\begin{document}
\title{Deep Learning and Open Set Malware Classification: A Survey}

\author{Jingyun Jia}
\affiliation{%
  \institution{School of Computing, Florida Institute of Technology}
  \streetaddress{150 West University Blvd}
  \city{Melbourne, FL 32901}
}
\email{jiaj2018@my.fit.edu}


\begin{abstract}
As the Internet is growing rapidly these years, the variant of malicious software, which often referred to as malware, has become one of the major and serious threats to Internet users. The dramatic increase of malware has led to a research area of not only using cutting edge machine learning techniques classify malware into their known families, moreover, recognize the unknown ones, which can be related to Open Set Recognition (OSR) problem in machine learning. Recent machine learning works have shed light on Open Set Recognition (OSR) from different scenarios. Under the situation of missing unknown training samples, the OSR system should not only correctly classify the known classes, but also recognize the unknown class. This survey provides an overview of different deep learning techniques, a discussion of OSR and graph representation solutions and an introduction of malware classification systems. 
\end{abstract}

%
%



\maketitle

\section{Introduction}

Malware, software that "deliberately fulfills the harmful intent of an attacker" \cite{bayer2006dynamic} nowadays come in a wide range of variations and multiple families. They have become one of the most terrible and major security threats of the Internet today. Instead of using traditional defenses, which typically use signature-based methods. There is an active research area that using machine learning-based techniques to solve the problem in both sides: 

\begin{enumerate}
    \item Classify known malware into their families, which turns out to be normal multi-classification
    \item Recognize unknown malware. i.e. malware that are not present in the training set but appear in the test set. 
\end{enumerate}

One solution for malware classification is to convert it to function call graphs (FCG), then classify them into families according to the representation of FCG as in \cite{hassen2017scalable}. Graphs are an important data structure in machine learning tasks, and the challenge is to find a way to represent graphs. Traditionally, the feature extraction relied on user-defined heuristics. And recent researches have been focus on using deep learning to automatically learn to encode graph structure into low-dimensional embedding. 

Another problem is it is less likely to label all the classes in training samples for the fast-developing diverse of malware families, the second item has become even more important daily. The related open set recognition (OSR) should also be able to handle those unlabeled ones. Traditional classification techniques focus on problems with labeled classes. While OSR pays attention to unknown classes. It requires the classifier accurately classify known classes, meanwhile identify unknown classes. Meanwhile, deep learning based OSR solutions have become a flourish research area in recent years. 

In this survey, we will first review basic deep learning techniques. Then
a brief categorization of current OSR techniques will be given in section 3. In section 4, we will cover methods learning graph representations, followed by an introduction to state-of-art malware classification techniques in section 5. And finally, section 6 will conclude.
\section{Deep learning basics}

As conventional machine-learning techniques were limited in their ability to process natural data in their raw form, hence required expertise in feature engineering \cite{lecun2015deep}. Deep Learning was introduced to discover intricate structures in high-dimension data, which requires litter engineering by hand.

In the following subsections, we will give an overview of different network architectures in different areas in recent years.

\begin{figure*}
\centering
\includegraphics[width=5in]{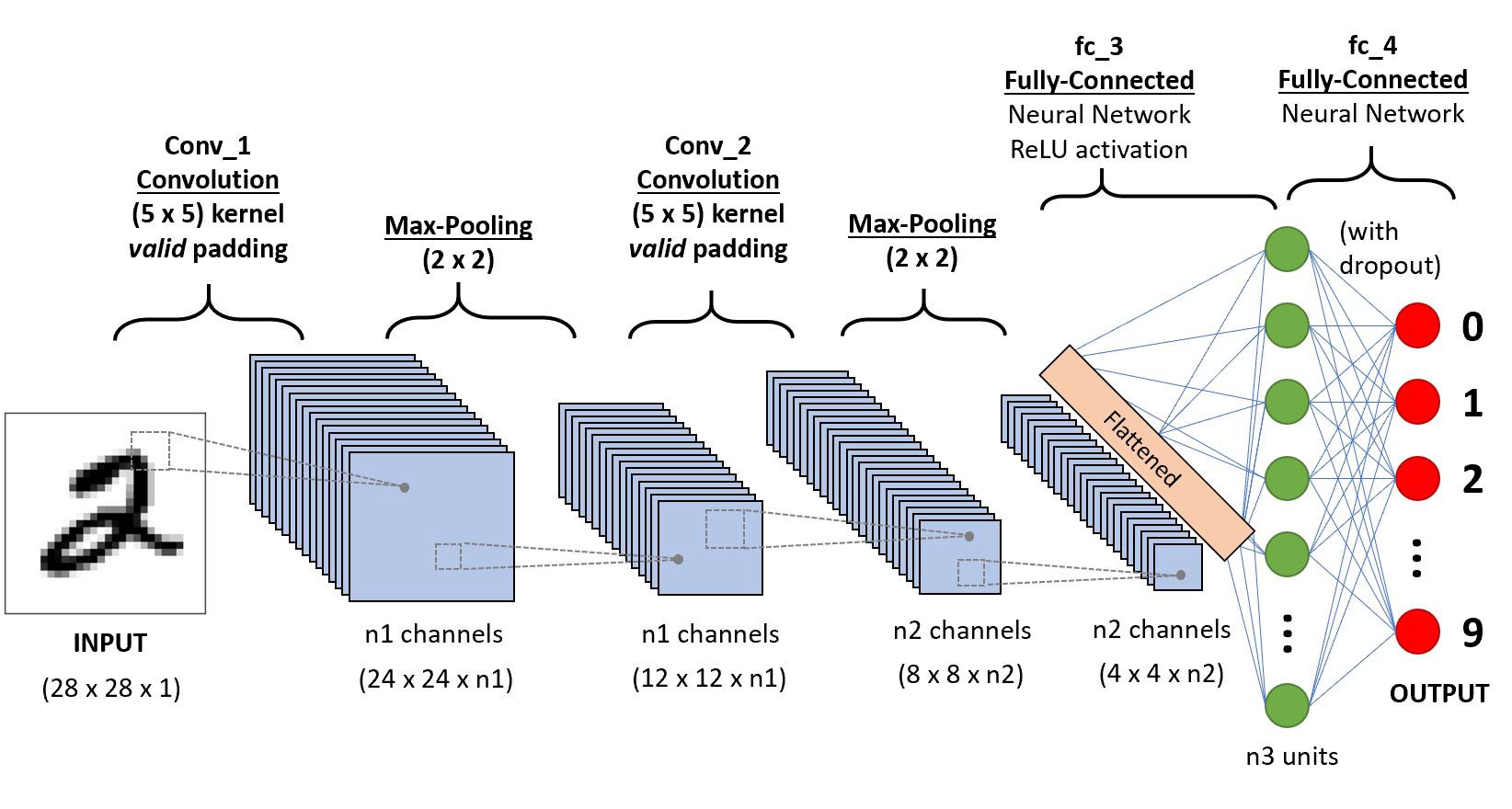}\Description{CNNs}
\caption{A CNN sequence to classify handwritten digits (\citeauthor{saha_2018}, \citeyear{saha_2018})}
\label{fig: cnns}
\end{figure*}

\subsection{Convolutional neural networks}

As a popular architecture of Deep Neural Networks, Convolutional Neural Networks (CNNs) has achieved good performance in the Computer Vision area. As Figure \ref{fig: cnns}, a typical CNN usually consists of “Input Layer”, “Convolutional Layer”, “Pooling Layer” and “Output Layer”. The convolutional and pooling layer can be repeated several times. In most cases, the Relu function is used as activation in the convolutional layer and Max-Pooling is used in the pooling layer. During the learning process, filters will be learned and feature maps will then be generated, which is the output of representation learning. The output is usually followed by a fully connected network for classification problems. The architectures could be different in various aspects: layers and connections (\cite{cao2015look} \cite{he2016deep}), loss functions (\cite{wen2016discriminative} \cite{he2018triplet} \cite{deng2019arcface}), etc.

\subsubsection{Feedback Network}
\citet{cao2015look} proposed Feedback Network to develop a computational feedback mechanism which can help better visualized and understand how deep neural network works, and capture visual attention on expected objects, even in images with cluttered background and multiple objects. 

Feedback Network introduced a feedback layer. The feedback layer contains another set of binary neuron activation variables  $Z\in \{0,1\}$. The feedback layer is stacked upon each ReLU layer, and they compose a hybrid control unit to active neuron response in both bottom-up and top-down manners: Bottom-Up Inherent the selectivity from ReLU layers, and the dominant features will be passed to upper layers; Top-Down is controlled by Feedback Layers, which propagate the high-level semantics and global information back to image representations. Only those gates related to particular target neurons are activated.


\subsubsection{Center loss}
Wen et al. proposed center loss as a new supervision signal (objective function) for face recognition tasks in \cite{wen2016discriminative}. To separate the features of different classes to achieve better performance in classification tasks, center loss tries to minimize the variation of intra-class. Let $\mathbf{c}_{y_i}$ denotes the center of the embeddings of $y_i$th class, the loss function looks like:

\begin{equation}
    \mathcal{L}_C = \frac{1}{2} \sum_{i=1}^m \| \mathbf{x}_i - \mathbf{c}_{y_i} \|_2^2,
\end{equation}

To make the computation more efficient, center loss uses a mini-batch updating method and the centers are updated by the features mean of the corresponding classes after each iteration. The paper showed that under the joint supervision of softmax loss and center loss, CNN can obtain inter-class dispensation and intra-class compactness as much as possible. 

\subsubsection{Triplet-center loss}
Inspired by triplet loss and center loss, He et al. introduced triplet-center loss to further enhance the discriminative power of the features, as to 2D object recognition algorithms in \cite{he2018triplet}. Triplet-loss intends to find an embedding space where the distances between different classes are greater than those form the same classes. Center loss tries to find embedding spaces where the deep learned features from the same class more compact and closer to the corresponding center. Similarly, instead of comparing the distances of each two instances in triplet loss, triplet loss computes the distances of instance and class center.

\begin{equation}
    \mathcal{L}_{tc}= \sum_{i=1}^M \max \left(D(f_i, C_{y_i}) + m - \min_{j \neq y_i} D(f_i, c_j), 0\right),
\end{equation}
where $D()$ is a distance function and $m$ is margin value. By setting up a margin value, the loss function ensures different classes be pushed by at least $m$ distance away.

\subsubsection{Arcface}
In \cite{deng2019arcface}, Deng et al. proposed an additive angular margin loss (ArcFace) to obtain highly discriminative features for face recognition. Based on classic softmax loss,

\begin{equation}
    \mathcal{L}_\text{softmax} = - \frac{1}{N} \sum_{i=1}^N \log \frac{e^{W_{y_i}^Tx_i+b_{y_i}}}{\sum_{j=1}^n e^{W_j^T x_i+b_j}},
\end{equation}
where $x_i$ denotes the embedding of the $i$th sample. After normalizing $x$ and $W$, ArcFace adds an additive angular margin penalty $m$ between $x_i$ and $W_{y_i}$ to simultaneously enhance the intra-class compactness and inter-class discrepancy:
\begin{equation}
    \mathcal{L}_\text{Arc} = - \frac{1}{N} \sum_{i=1}^N \log \frac{e^{s(\cos(\theta_{y_i}+m))}}{e^{s(\cos(\theta_{y_i}+m))} + \sum_{j=1, j \neq y_i}^n e^{s \cos \theta_j}}
\end{equation}

The paper shows that ArcFace has a better geometric attribute as the angular margin has the exact correspondence to the geodesic distance.

\subsubsection{ResNets}
\begin{figure}
\centering
\includegraphics[width=3in]{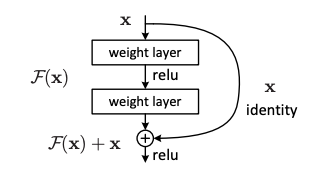}\Description{resNet}
\caption{A building block of resNet (\citeauthor{he2016deep}, \citeyear{he2016deep})}
\label{fig: resNet}
\end{figure}

The training of deeper neural networks is facing degradation problems: with the network depth increasing, accuracy gets saturated and then degrades rapidly. The degradation problem indicates that not all systems are similarly easy to optimize. Under the hypothesis that it is easier to optimize the residual mapping than to optimize the original unreferenced mapping, \citet{he2016deep} presented a residual learning framework called ResNets to ease the training of the network. ResNets consists of residual blocks as Figure \ref{fig: resNet}. Instead of hoping each few stacked layers directly fit a desired underlying mapping, ResNets explicitly let these layers fit a residual mapping.  

Specifically, instead fitting the desired underlying mapping as $H(x)$, ResNet makes stacked nonlinear layers fit another mapping of $\mathcal{F}(x) := H(x) - x$. Then original mapping is recast into $\mathcal{F}(x) + x$. In an extreme case, if an identity mapping were optimal, it would be easier to push the residual to zero than to fit an identity mapping.

\begin{figure}
\centering
\includegraphics[width=3in]{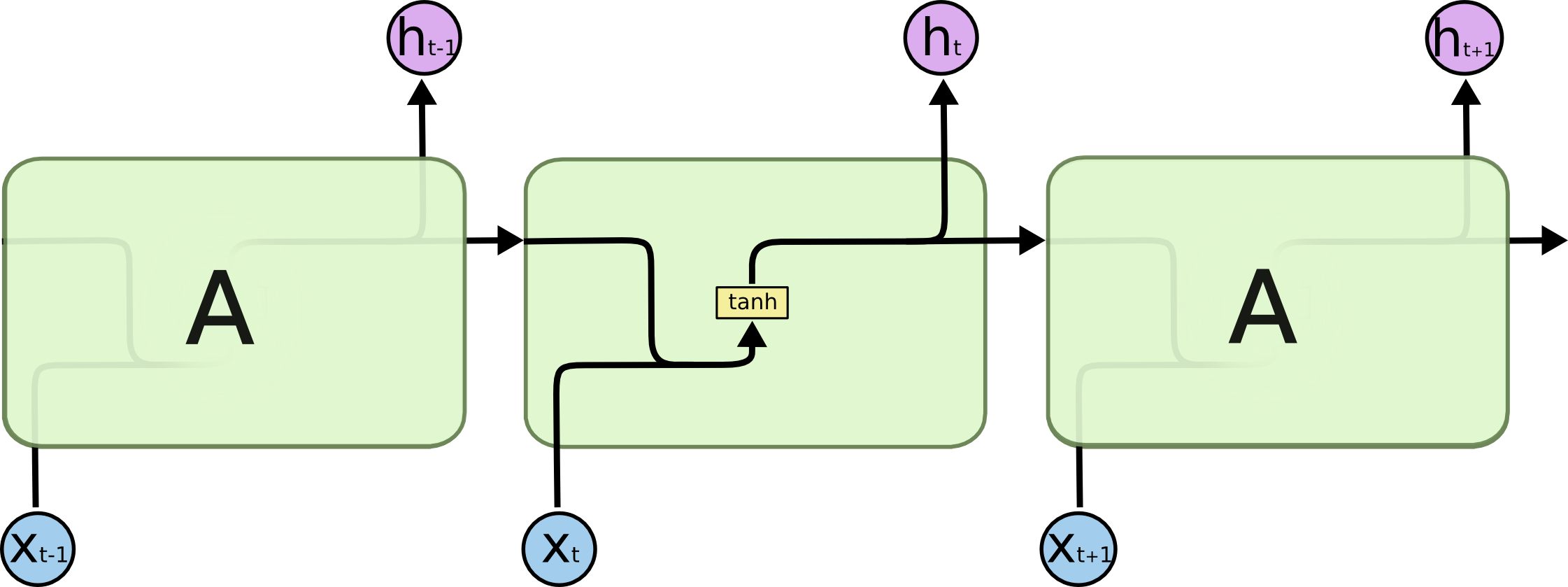}\Description{RNNs}
\caption{A standard RNN contains a single layer (\citeauthor{olah_2015}, \citeyear{olah_2015})}
\label{fig: RNN}
\end{figure}

\subsection{Recurrent neural networks}
Another type of popular architecture of Deep Neural Networks is Recurrent Neural Networks (RNNs), which involves sequential inputs, such as speech and language. RNNs process an input sequence one element at a time, also maintain a state vector that implicitly contains all the historical information. An unfold RNN (Figure \ref{fig: RNN}) could be considered as a deep multi-layer network. Just like CNNs, there are multiple variants for RNNs as well. Particularly, it has been widely used in machine translation tasks (\cite{DBLP:conf/emnlp/ChoMGBBSB14} \cite{DBLP:journals/corr/BahdanauCB14} \cite{DBLP:conf/emnlp/LuongPM15}).

\begin{figure}
\centering
\includegraphics[width=3in]{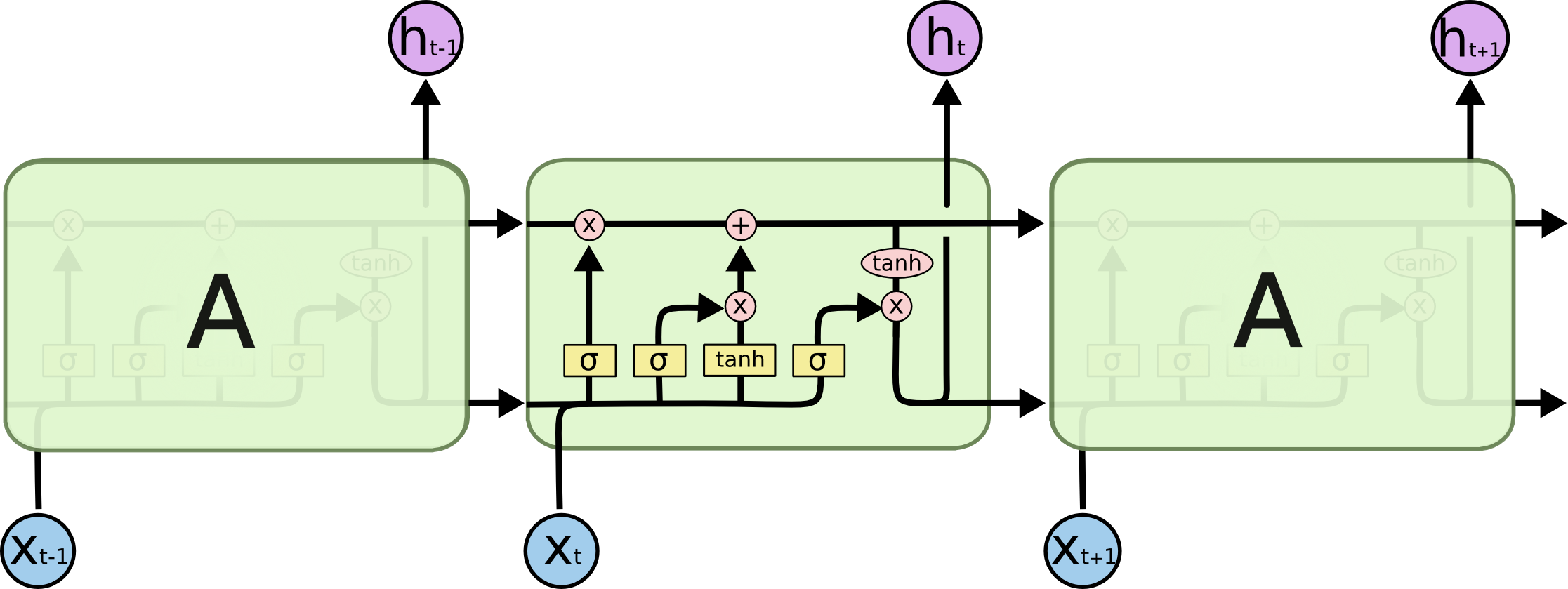}\Description{LSTMs}
\caption{An LSTM contains four interacting layers (\citeauthor{olah_2015}, \citeyear{olah_2015})}
\label{fig: lstm}
\end{figure}

\subsubsection{LSTM}
As it is not applicable to store information for very long, “Long Short Term Memory” (LSTM) was proposed to solve the problem. \citeauthor{chen2016gentle} gave a gentle tutorial on basics of backpropagation in recurrent neural networks (RNN) and long short-term memory (LSTM) in \cite{chen2016gentle}. LSTM (Figure \ref{fig: lstm}) includes four gates: input modulation gate, input gate, forget gate (\citet{gers1999learning}) and output gate along with their corresponding weights. LSTM also contains a special unit called memory cell act like an accumulator or a gated leaky neuron. Meanwhile, There are other augment RNNs with a memory module such as “Neural Turing Machine” and “memory networks”. These models are being used for tasks need reasoning and symbol manipulation.

\begin{figure}
\centering
\includegraphics[width=2in]{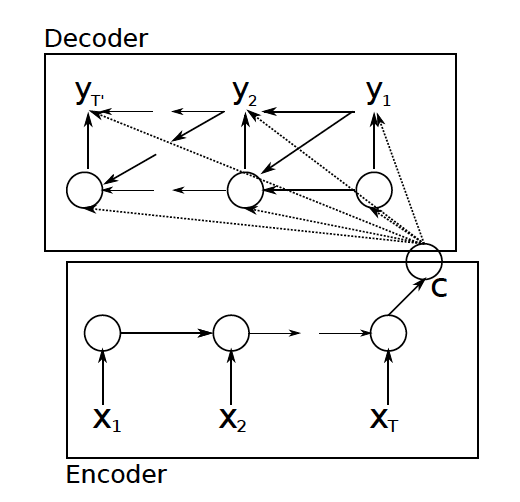}\Description{rnn_en_de}
\caption{An illustration of the RNN Encoder-Decoder (\citeauthor{DBLP:conf/emnlp/ChoMGBBSB14}, \citeyear{DBLP:conf/emnlp/ChoMGBBSB14})}
\label{fig: rnn_en_de}
\end{figure}

\subsubsection{RNN encoder-dencoder}
\citet{DBLP:conf/emnlp/ChoMGBBSB14} proposed a neural network architecture called RNN Encoder-Decoder (Figure \ref{fig: rnn_en_de}), which can be to used as additional features in statistical machine translation (SMT) system to generate a target sequence, also can be used to score a given pair of input and output sequence. The architecture learns to encode a variable-length sequence into a fixed-length vector representation and to decode a given fixed-length vector representation back into a variable-length sequence.
The encoder is an RNN that reads each symbol of an input sequence x sequentially.
The decoder of the proposed model is another RNN that is trained to generate the output sequence by predicting the next symbol given the hidden state. 

In addition to a novel model architecture, the paper also proposed a variant of LSTM, which includes an update gate and a reset gate. The update gate selects whether the hidden state is to be updated with a new hidden state while the reset gate decides whether the previous hidden state is ignored.

\begin{figure}
\centering
\includegraphics[width=1.5in]{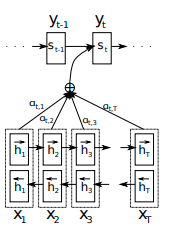}\Description{rnnsearch}
\caption{An illustration of the RNNsearch (\citeauthor{DBLP:journals/corr/BahdanauCB14}, \citeyear{DBLP:journals/corr/BahdanauCB14})}
\label{fig: rnnsearch}
\end{figure}

\subsubsection{RNNsearch}
In \cite{DBLP:journals/corr/BahdanauCB14}, \citeauthor{DBLP:journals/corr/BahdanauCB14} proposed a new architecture for machine translation model by adding an alignment model to basic RNN Encoder-Decoder. Just like a traditional machine translation model, the proposed architecture consists of an encoder and a decoder. The encoder reads the input sentence, then converts into a vector. And the decoder emulates searching through a source sentence during decoding a translation. As Figure \ref{fig: rnnsearch}, the align model learns the weights $\alpha_{ij}$ of each annotation $h_j$ scoring how well the inputs around the position $j$  and output at position $i$ match. The score is based on the RNN hidden state $s_{i-1}$ and the $j$th annotation $h_j$ of the input sentence.

\subsubsection{Attentional mechanism}
In \cite{DBLP:conf/emnlp/LuongPM15}, \citeauthor{DBLP:conf/emnlp/LuongPM15} examined two classes of attentional mechanism to better improve neural machine translation (NMT): a global approach which always attends to all source words and a local one that only looks at a subset of source words at a time. Based on LSTM, they introduced a variable-length alignment vector for two kinds of attentional mechanisms. The global attention model is based on the global context, and the size of the alignment vector equals the number of time steps on the source site. While the local attention model is based on a window context, where the size of the alignment vector equals to window size.

\begin{figure}
\centering
\includegraphics[width=3in]{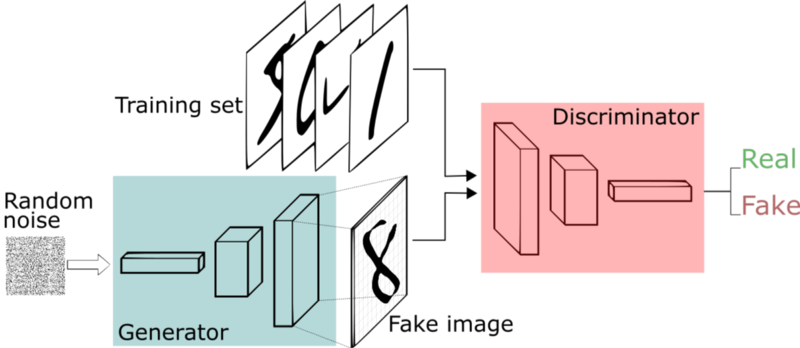}\Description{GANs}
\caption{Overview of the framework of GANs (\citeauthor{silva_2018}, \citeyear{silva_2018})}
\label{fig: gans}
\end{figure}

\subsection{Generative adversarial networks}
Deep learning has achieved great performance in supervised learning in discriminative models. However, deep generative models have had less of impact as:
\begin{itemize}
    \item It is difficult to approximate the computations in maximum likelihood estimation
    \item It is difficult to leverage the benefits of piecewise linear units in the generative context”
\end{itemize}

\citet{goodfellow2014generative} proposed a new generative model: generative adversarial nets (GANs) to avoid these difficulties. The proposed GANs architecture includes two components: a generator $G$ and a discriminator $D$. To learn the distribution $P_g$ over given data $x$. GANs define a prior on input noise variables $p_\mathbf{z}(\mathbf{z})$. The framework of GANs corresponds to a “min-max two-player game” (discriminator vs. generator) with value function $V(G, D)$: 


\begin{equation}
    \min_G \max_D V(D, G) = \mathbb{E}_{\mathbf{x}\sim p_{\text{data}}(\mathbf{x})}[\log D(\mathbf{x})] + \mathbb{E}_{\mathbf{z} \sim p_\mathbf{z} (\mathbf{z})} [\log(1 - D(G(\mathbf{z})))]
\end{equation}

Generator generates noise samples from a prior distribution and discriminator represents the probability of the data come from the target dataset rather than a generator. Hence the target is to train the discriminator $D$ to maximize the probability of assigning the correct label to both training examples and samples from $G$, meanwhile train the generator $G$ to minimize $\log(1 - D(G(\mathbf{z})))$, i.e. generating samples alike examples to ``fool'' the discriminator. In practice, the procedure optimizes $D$ $k$ steps and one step of $G$.

\subsubsection{DCGANs}

\citet{DBLP:journals/corr/RadfordMC15}  proposed deep convolution generative adversarial networks (DCGANs) to bridge the gap between the supervised learning and unsupervised learning in CNNs, which makes GANs more stable. The architecture guidelines for stable Deep Convolutional GANs:

\begin{itemize}
    \item Replace any pooling layers with stridden convolutions (discriminator) and fractional-strided convolutions (generator).
    \item Use batch norm in both the generator and the discriminator.
    \item Remove fully connected hidden layers for deeper architectures.
    \item Use ReLU activation in the generator for all layers except for the output, which uses Tanh.
    \item Use LeakyReLU activation in the discriminator for all layers.
\end{itemize}

\subsubsection{AAE}
\citet{makhzani2015adversarial} proposed a new inference algorithm Adversarial Autoencoder (AAE), which uses the GANs framework which could better deal with applications such as semi-supervised classification, disentangling style and content of images, unsupervised clustering, dimensionality reduction, and data visualization. The algorithm aims to find a representation for graphs that follows a certain type of distribution. And it consists of two phases: the reconstruction phase and the regularization phase. In the reconstruction phase, encoder and decoder are updated to minimize reconstruction error. In the regularization phase, the discriminator is updated to distinguish true prior samples from generated samples, and the generator is updated to fool the discriminator. Reconstruction phase and regularization phase are referred to as the generator and discriminator in GANs. And the method could be used in semi-supervised learning and unsupervised clustering. For semi-supervised learning, there is a semi-supervised classification phase besides the reconstruction phase and regularization phase. And labeled data would be trained at this stage. which is an aggregated categorical distribution. The architecture of unsupervised clustering is similar to semi-supervised learning, the difference is that the semi-supervised classification stage is removed and thus no longer train the network on any labeled mini-batch.

\subsection{Representation learning}

Representation learning allows a machine to be fed with raw data and to automatically discover the representations (embeddings) needed for detection or classification \cite{lecun2015deep}. Those raw data could be images, videos, texts, etc. An image comes in the form of an array of pixel values and texts come in the form of word sequences. Motivated by different objectives, a set of representative features would be generated through deep neural networks.

\subsubsection{Skip-gram}
Skip-gram model has achieved good performance in learning high-quality vector representations of words from large amounts of unstructured text data, which doesn’t require dense matrix multiplications. The training objective of the Skip-gram model is to find word representations that are useful for
predicting the surrounding words in a sentence or a document. Given a sequence of training words $\mathfrak{w}_1, \mathfrak{w}_2, \mathfrak{w}_3,..., \mathfrak{w}_T$, the objective of the Skip-gram model is to maximize the average log probability:

\begin{equation}
    \frac{1}{T} \sum_{t=1}^T \sum_{-c \leq j \leq c, j \neq 0} \log p(\mathfrak{w}_{t+j} | \mathfrak{w}_t),
\end{equation}
where $c$ is the size of training context. The basic Skip-gram formulation defines $p(\mathfrak{w}_{t+j} | \mathfrak{w}_t)$ using the softmax function:

\begin{equation}
    p(\mathfrak{w}_\mathit{O} | \mathfrak{w}_\mathit{I}) = \frac{\exp (
    \mathfrak{v}'^\top_{\mathfrak{w}_\mathit{O}} 
    \mathfrak{v}_{\mathfrak{w}_\mathit{I}})}
    {\sum_{\mathfrak{w}=1}^W \exp (
    \mathfrak{v}'^\top_\mathfrak{w} \mathfrak{v}_{\mathfrak{w}_\mathit{I}})}.
\end{equation}
where $\mathfrak{v}_\mathfrak{w}$ and $\mathfrak{v}'_\mathfrak{w}$ are the "input" and "output" vector representations of $\mathfrak{w}$, and $W$ is the number of words in the vocabulary. Based on the skip-gram algorithm. \citet{mikolov2013distributed} presents some extensions to improve its performance: Hierarchical softmax, negative sampling and subsampling. It shows that the word vectors can be meaningful combined using just simple vector addition. Specifically, hierarchical softmax uses a binary tree to present the output layer rather than a plat output of all the words for output dimension reduction to make the computation more efficient. An alternative to hierarchical softmax is negative sampling, inspired by Noise Contrastive Estimation (NCE). The basic idea is to sample one “accurate” data and $k$ noise data, the objective is to maximize their conditional log-likelihood:

\begin{equation}
    \log \sigma(\mathfrak{v}'^\top_{\mathfrak{w}_\mathit{O}} \mathfrak{v}_{\mathfrak{w}_\mathit{I}}) +
    \sum_{i=1}^k \mathbb{E}_{\mathfrak{w}_i \sim P_n(\mathfrak{w})}\left[ \log \sigma(-v'^\top_{\mathfrak{w}_i} v_{\mathfrak{w}_\mathit{I}}) \right],
\end{equation}

The objective of NCE is used to replace every $\log P(\mathfrak{w}_\mathit{O} | \mathfrak{w}_\mathit{I})$ term in the Skip-gram objective, and the task is to distinguish the target word $\mathfrak{w}_\mathit{O}$ from draws from the noise distribution $P_n(\mathfrak{w})$ using logistic regression, where there are $k$ negative samples for each data sample. The paper also suggested a simple subsampling approach to address the imbalance issue between the rare and frequent words: each word $w_i$ in the training set is discarded with probability computed by the formula
\begin{equation}
    P(w_i) = 1 - \sqrt{\frac{t}{f(w_i)}},
\end{equation}
where $f(w_i)$ is the frequent of word $w_i$ and $t$ is threshold.

\subsubsection{GloVe}
To better deal with word representations: word analogy, word similarity, and named entity recognition tasks, \citet{pennington2014glove} constructs a new model GloVe (for Global Vectors), which can capture the global corpus statistics. GloVe combines count-based methods and prediction-based methods for the unsupervised learning of word representations, proposing a new cost function
\begin{equation}
    J = \sum_{i, j=1}^V f(X_{ij})(W_i^T\Tilde{w}_j + b_i + \Tilde{b}_j - \log X_{ij})^2,
\end{equation}
where $V$ is the size of the vocabulary, $f(X_{ij})$ is a weighting function. $w$ are word vectors and $\Tilde{w}$ are separate context word vector as training multiple instances of the network and then combining the results can help reduce overfitting and noise and generally improve results. $b$ and $\Tilde{b}$ are corresponding bias for $w$ and $\Tilde{w}$.

\subsection{Meta-learning and interpretability}

\begin{figure}
\centering
\includegraphics[width=2in]{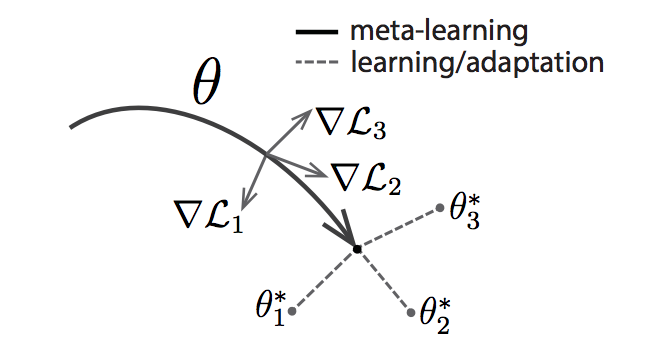}\Description{MAML}
\caption{Diagram of the MAML (\citeauthor{finn2017model}, \citeyear{finn2017model})}
\label{fig: MAML}
\end{figure}

\subsubsection{Meta-Learning}
Deep neural networks generally perform poorly on few-shot learning tasks as a classifier has to quickly generalize after seeing very few examples from each class. \citet{ravi2016optimization} proposes an LSTM based meta-learner model to learn the exact optimization algorithm used to train another learner neural network classifier in the few-shot regime. The meta-learner captures both short-term knowledge within a task and
long-term knowledge common among all the tasks. Also, \citet{finn2017model} proposed an algorithm called model-agnostic meta-learning (MAML) for meta-learning which is compatible with any model trained with gradient descent and different learning problems such as classification, regression and reinforcement learning. The meta-learning is to prepare the model for fast adaption. In general, it consists of two steps:

\begin{enumerate}
    \item sample batch of tasks to learn the gradient update for each of them, then combine their results;
    \item take the result of step 1 as a starting point when learning a specific task.
\end{enumerate}

The diagram looks like Figure \ref{fig: MAML}, it optimizes for a representation $\theta$ that can quickly adapt to new tasks.

\subsubsection{LIME}
As machine learning techniques are rapidly developed these days, there are plenty of models remain mostly black box. To make the predictions more interpretable to non-expertises despite which model it is (model-agnostic), so that people can make better decisions, \citet{ribeiro2016should} proposed a method - Local Interpretable Model-agnostic Explanations (LIME) to identify an interpretable model over interpretable presentation that is locally faithful to the classifier. It introduced sparse line explanations, weighing similarity between instance and its interpretable version with their distance. The paper also suggested a submodular pick algorithm (SP-LIME) to better select instances by picking the most important features based on the explanation matrix learned in LIME.

\subsubsection{Large-Scale evolution}
To minimize human participation in neural network design, Real et al. \cite{real2017large} employed evolutionary algorithms to discover network architectures automatically. The evolutionary algorithm uses the evolutionary algorithm to select the best of a pair to be a parent during tournament selection. Using pairwise comparisons instead of whole population operations. 

In the proposed method, individual architectures are encoded as a graph.  In the graph, the vertices represent
rank-3 tensors or activations. The graph’s edges represent identity connections or convolutions and contain the mutable numerical parameters defining the convolution’s properties. A child is similar but not identical to the parent because of the action of a mutation. Mutation operations include "ALTER-LEARNING-RATE", "RESET-WEIGHTS", "INSERT-CONVOLUTION", "REMOVE-CONVOLUTION", etc.

\section{Open Set Recognition}
As deep learning has achieved great success in object classification, it is less likely to label all the classes in training samples, and the unlabeled class, so-called open-set becomes a problem. Different from a traditional close-set problem, which only requires correctly classify the labeled data, open set recognition (OSR) should also handle those unlabeled ones. \citeauthor{geng2018recent} stated four categories of recognition problems as follows \cite{geng2018recent}:

\begin{enumerate}
\item known known classes: labeled distinctive positive classes, available in training samples;
\item known unknown classes: labeled negative classes, available in training samples;
\item unknown known classes: training samples not available, but having some side-information such as semantic/attribute information
\item unknown unknown classes: neither training samples nor side-information available, completely unseen.
\end{enumerate}

Traditional classification techniques focus on problems with labeled classes, which include known known classes and known unknown classes. While open set recognition (OSR) pays attention to the later ones: unknown known classes and unknown unknown classes. It requires the classifier accurately classify known known classes, meanwhile identify unknown unknown classes.

In general, the techniques can be categorized into three classes according to the training set compositions as Table \ref{tab:cat}.                                                          

\begin{table}
  \caption{OSR Techniques Categorization}
  \label{tab:cat}
  \begin{tabular}{p{4.1cm}|p{3.6cm}}
    \toprule
    Training Set & papers\\
    \midrule
    \texttt{Borrowing Additional Data}&
\cite{shu2018unseen}
\cite{saito2018open}
\cite{shu2018odn}
\cite{DBLP:conf/iclr/HendrycksMD19}
\cite{dhamija2018reducing}
\cite{perera2019learning}\\

    \texttt{Generating additional data}& 
\cite{jo2018open}
\cite{neal2018open}
\cite{DBLP:conf/bmvc/GeDG17}
\cite{DBLP:conf/ijcai/YuQLG17}
\cite{DBLP:conf/iclr/LeeLLS18} \\

    \texttt{No Additional Data} & 
\cite{bendale2016towards}
\cite{hassen2018learning}
\cite{junior2017nearest}
\cite{mao2018distribution}
\cite{wang2018co}
\cite{schultheiss2017finding}
\cite{zhang2016sparse}
\cite{DBLP:conf/iclr/LiangLS18}
\cite{DBLP:conf/emnlp/ShuXL17}\\
    
    \bottomrule
  \end{tabular}
\end{table}

\subsection{Borrowing additional data}

To better discriminate known class and unknown class, some techniques introduce unlabeled data in training(\cite{shu2018unseen} \cite{saito2018open}). In addition, \cite{shu2018odn} indicates several manually annotations for unknown classes are required in their workflow.

\subsubsection{Open set domain adaptation by backpropagation}

\citet{saito2018open} proposed a method which marks unlabeled target samples as unknown, then mixes them with labeled source samples together to train a feature generator and a classifier. The classifier attempts to make a boundary between source and target samples whereas the generator attempts to make target samples far from the boundary. The idea is to extract feature which separates known and unknown samples. According to the feature generator, the test data either would be aligned to known classes or rejected as an unknown class. 

\subsubsection{Unseen class discovery in open-world classification}

\begin{figure*}
\includegraphics[height=2in, width=6in]{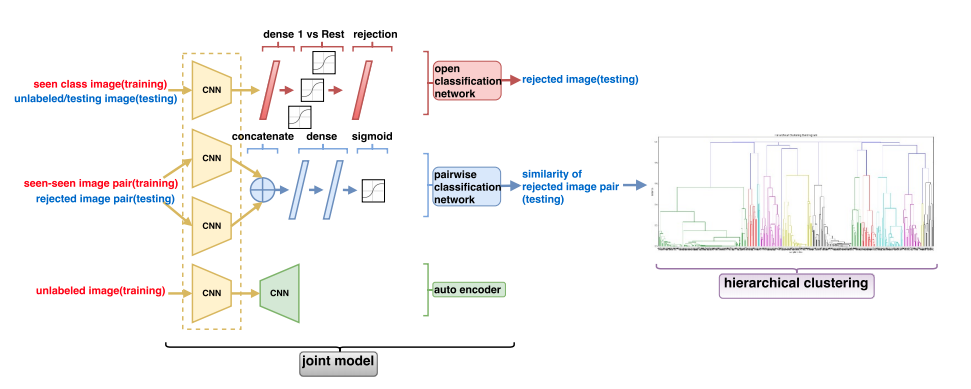}\Description{OCN+PCN+HC}
\caption{Overall framework of OCN+PCN+HC (\citeauthor{shu2018unseen}, 2018)}
\label{fig:unseen}
\end{figure*}

\citet{shu2018unseen} introduced a framework to solve the open set problem, which involves unlabeled data as an autoencoder network to avoid overfitting. Besides autoencoder, it contains another two networks in the training process - an Open Classification Network (OCN), a Pairwise Classification Network (PCN). Only OCN participants in the testing phase, which predicts test dataset including unlabeled examples from both seen and unseen classes. Then it follows the clustering phase, based on the results of the predictions of PCN, they used hierarchical clustering (bottom-up/ merge) to cluster rejected examples clusters. Overall framework as Figure \ref{fig:unseen}.

\subsubsection{ODN}
Manual labeled unknown data is used in Open Deep Network (ODN) proposed by \citet{shu2018odn}. It needs several manually annotations. Specifically, it added another new column corresponding to the unknown category to the weight matrix and initialized it as $\mathfrak{w}_{N+1}$:
\begin{equation}
    \mathfrak{w}_{N+1} = \alpha \frac{1}{N} \sum_{n=1}^N \mathfrak{w}_n + \beta \frac{1}{M} \sum_{m=1}^M \mathfrak{w}_m,
\end{equation}
where $\mathfrak{w}_n$ is the weight column if the known $n$th category. In addition, as the similar categories should play a more critical role in the initialization of $\mathfrak{w}_m$, ODN added another term $\frac{1}{M}\sum_{m=1}^M \mathfrak{w}_m$ to emphase the similar known categories. The $\mathfrak{w}_m$ is the weight columns of $M$ highest activation values. The $\mathfrak{w}_{N+1}$ is concatenated to the transfer weight $W$ to support the new category. And this initialization method is called Emphasis Initialization.

ODN also introduces multi-class triplet thresholds to identify new categories: accept threshold, reject threshold and distance-threshold. Specifically, a sample would be accepted as a labeled class if and only if the index of its top confidence value is greater than the acceptable threshold. A sample would be considered as unknown if all the confidence values are below the rejected threshold. For samples between accept threshold and reject threshold, they would also be accepted as a labeled class if the distance between top and second maximal confidence values is large than the distance-threshold.

\subsubsection{Outlier Exposure}
\citet{DBLP:conf/iclr/HendrycksMD19} proposed Outlier Exposure(OE) to distinguish between anomalous and in-distribution examples. OE borrowed data from other datasets to be ``out-of-distribution'' (OOD), denoted as $\mathcal{D}_{out}$. Meanwhile target samples as ``in-distribution'', marked as $\mathcal{D}_{in}$. Then the model is trained to discover signals and learn heuristics to detect” which dataset a query is sampled from. Given a model $f$ and the original learning objective $\mathcal{L}$, the objective function of OE looks like:

\begin{equation} 
    \mathbb{E}_{(x,y) \sim \mathcal{D}_{in}}[\mathcal{L}(f(x),y) + \lambda \mathbb{E}_{x' \sim \mathcal{D}_{in}^{out}}[\mathcal{L}_{OE}(f(x'), f(x), y)]]
\end{equation}

$\mathcal{D}_{out}^{OE}$ is an outlier exposure dataset. The equation indicates the model tries to minimize the objective L for data from ``in-distribution'' ($\mathcal{L}$) and ``out-of-distribution'' ($\mathcal{L}_{OE}$). The paper also used the maximum softmax probability baseline detector (cross-entropy) for $\mathcal{L}_{OE}$. And when labels are not available, $\mathcal{L}_{OE}$ was set to a margin ranking loss on the log probabilities $f(x')$ and $f(x)$. However, the performance of this method depends on the chosen OOD dataset.

\subsubsection{Objectosphere Loss}
\citet{dhamija2018reducing} proposed Entropic Open-Set and Objectoshere losses for open set recognition, which trained networks using negative samples from some classes. The method reduced the deep feature magnitude and maximize entropy of the softmax scores of unknown sample to separate them from known samples. The idea of Entropic Open-Set is to maximum entropy when an input is unknown. Formally, let $S_c(x)$ denotes the softmax score of sample $x$ from known class $c$, the Entropic Open-Set Loss $J_E$ can be defined as:
\begin{equation}
J_E(x) = 
    \begin{cases} 
    -\log S_c(x) & \text{if $x\in \mathcal{D}'_c$ is from class $c$} \\
    - \frac{1}{C} \sum_{c=1}^C \log S_c(x) & \text{if $x\in \mathcal{D}'_b$ is from class $c$}
    \end{cases}
\end{equation}
where $\mathcal{D}'_b$ denotes out of distribution samples. To further separate known and unknown samples, the paper pushed known samples into the “objectosphere” where they have large feature magnitude and low entropy, so-called Objectosphere Loss, which is calculated as:
\begin{equation}
J_R = J_E + \lambda
    \begin{cases} 
    \max(\xi - \|F(x)\|, 0)^2 & \text{if $x \in \mathcal{D}'_c$} \\
    \|F(x)\|^2 & \text{if $x \in \mathcal{D}'_b$}
    \end{cases}
\end{equation}
where $F(x)$ is the deep feature vector, and Objectosphere Loss penalizes the known classes if their feature magnitude is inside the boundary of the Objectosphere $\xi$ and unknown classes if their magnitude is greater than zero.

\subsubsection{DOC}
\citet{perera2019learning} proposed a deep learning-based solution for one-class classification (DOC) feature extraction. The objective of one-class classification is to recognize normal class and abnormal class using only samples from normal class and there are different strategies to solve a classification problem. The proposed accept two inputs: one from the target dataset, one from the reference dataset, and produces two losses through a pre-trained reference network and a secondary network. The reference dataset is the dataset used to train the reference network, and the target dataset contains samples of the class for which one-class learning is used for. During training, two image batches, each from the reference dataset and the target dataset are simultaneously fed into the input layers of the reference network and secondary network. At the end of the forward pass, the reference network generates a descriptiveness loss ($l_D$), which is the same as the cross-entropy loss, and the secondary network generates compactness loss ($l_C$). The composite loss ($l$) of the network is defined as:
\begin{equation}
    l(r,t) = l_D(r|W) + \lambda l_C(t|W),
\end{equation}
where $r$ and $t$ are the training data from reference dataset and target dataset respectively, $W$ is the shared weights of both networks, and $\lambda$ is a constant.

\begin{figure}
\centering
\includegraphics[width=3in]{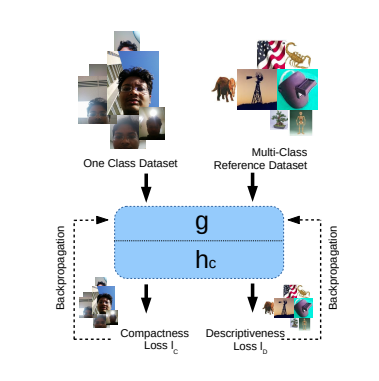}\Description{OC}
\caption{Overview of the Deep Feature for One-Class Classification framework (\citeauthor{perera2019learning}, \citeyear{perera2019learning})}
\label{fig: oc}
\end{figure}

The overview of the proposed method looks like figure \ref{fig: oc}, where $g$ is feature extraction networks and $h_c$ is classification networks. The compactness loss is to assess the compactness of the class under consideration in the learned feature space. The descriptiveness loss was assessed by an external multi-class dataset.

\bigbreak

\noindent\textbf{Discussion} \\
All the above methods borrow some dataset as unknown class during training, \cite{shu2018unseen} borrows target samples (test set) as unknown classes and utilizes adversarial learning. A classifier is trained to make a boundary between the source and the target samples whereas a generator is trained to make target samples far from the boundary. \cite{saito2018open} also borrows unlabeled examples from the dataset, then used and Auto-encoder to learn the representations. \cite{shu2018odn} uses several manually annotations during Emphasis Initialization. \cite{DBLP:conf/iclr/HendrycksMD19} introduced outlier exposure datasets on top of in-distribution datasets.  \cite{dhamija2018reducing} also introduces unknown datasets, meanwhile utilizes the differences of feature magnitudes between known and unknown samples as part of the objective function. Different from the multi-class classification problems, \cite{perera2019learning} presents a one-class classification problem from anomaly detection, with additional reference dataset for transfer learning. In general, borrowing and annotating additional data s OSR an easier problem. However, the retrieval and selection of additional datasets would be another problem.

\subsection{Generating additional data}
As adversarial learning has achieved great access such as GANs, there are ideas use GANs generating unknown samples before training. 

\subsubsection{G-OpenMax}
\citet{DBLP:conf/bmvc/GeDG17} designed a networks based on OpenMax and GANs. Their approach provided explicit modeling and decision score for novel category image synthesis. The method proposed has two stages as well as OpenMax: pre-Network training and score calibration. During the pre-Network training stage, different with OpenMax, it first generates some unknown class samples (synthetic samples) then sends them along with known samples into networks for training. A modified conditional GAN is employed in G-OpenMax to sythesize unknown classes. In conditional GAN, random noise is fed to the generator $G$ with a one-hot vector $c \in c_{i,...,k}$, which represents a desired class. Meanwhile, the discriminator $D$ learns faster if the input image is supplied together with the class it belongs to. Thus, the optimization of a conditional GAN with class labels can be formulated as:

\begin{equation}
\begin{aligned}
    \min_\phi \max_\theta & = E_{x,c\sim P_{data}} [\log D_\theta (x,c)]\\
                           & + E_{z \sim P_{z}, c \sim P_{c}}[\log (1-D_\theta(G_\phi (z,c),c))],
\end{aligned}
\end{equation}
where $\phi$ and $\theta$ are trainable parameters for $G_\phi$ and $D_\theta$, the generator inputs $z$ and $c$ are the latent variables drawn from their prior distribution $P(z)$ and $P(c)$. For each generated sample, if the class with the highest value is different from the pre-trained classifier, it will be marked as "unknown". Finally, a final classifier provides an explicit estimated probability for generated unknown classes.

\subsubsection{Adversarial sample generation}
\citet{DBLP:conf/ijcai/YuQLG17} proposed Adversarial Sample Generation (ASG) as a data augmentation technique for the OSR problem. The idea is to generate some points closed to but different from the training instances as unknown labels, then straightforward to train an open-category classifier to identify seen from unseen. Moreover, ASG also generates "unknown" samples, which are close to "known" samples. Different from the GANs min-max strategy, ASG generated samples based on distances and distributions, the generated unknown samples are: 
\begin{enumerate}
    \item close to the seen class data
    \item scattered around the known/unknown boundary
\end{enumerate}

\subsubsection{Counterfactual image generation}

Different from standard GANs, \citet{neal2018open} proposed a dataset augmentation technique called counterfactual image generation, which adopts an encoder-decoder architecture to generate synthetic images closed to the real image but not in any known classes, then take them as unknown class. The architecture consists of three components:
\begin{itemize}
    \item An encoder network: maps from images to a latent space.
    \item A generator: maps from latent space back to an image.
    \item A discriminator: discriminates generated images from real images.
\end{itemize}

\subsubsection{GAN-MDFM}
\citet{jo2018open} presented a new method to generate fake data for unknown unknowns. They proposed Marginal Denoising Autoencoder (MDAE) technique, which models the noise distribution of known classes in feature spaces with a margin is introduced to generate data similar to known classes but not the same ones. The model contains a classifier, a generator, and an autoencoder. The classifier calculated the entropy of membership probability instead of discriminating generated data from real data explicitly. Then, a threshold is used here to identify unknown classes. The generator modeled the distribution $m$ away from the known classes.

\subsubsection{Confident classifier} 
In order to distinguish 
in-distribution and out-of-distribution samples, \citet{DBLP:conf/iclr/LeeLLS18} suggested two additional terms added to the original cross-entropy loss, where the first one (confident loss) forces out-of-distribution samples less confident by the classifier while the second one (adversarial generator) is for generating most effective training samples for the first one. Specifically, the proposed confident loss is to minimize the Kullback-Leibler (KL) divergence from the predictive distribution on out-of-distribution samples to the uniform one in order to achieve less confident predictions for samples from out-of-distribution. Meanwhile in- and out-of-distributions are expected to be more separable. Then, an adversarial generator is introduced to generate the most effective out-of-distribution samples. Unlike the original generative adversarial network(GAN), which generates samples similar to in-distribution samples, the proposed generator generates "boundary" samples in the low-density area of in-distribution acting as out-of-distribution samples. Finally, a joint training scheme was designed to minimize both loss functions alternatively. Finally, the paper showed that the proposed GAN implicitly encourages training a more confident classifier.

\bigbreak

\noindent\textbf{Discussion} \\
Instead of borrowing data from other datasets, generating additional data methods generate unknown samples from the knowns. Most data generation methods are based on GANs. \cite{DBLP:conf/bmvc/GeDG17} introduces a conditional GAN to generate some unknown samples followed by OpenMax open set classifier. \cite{DBLP:conf/ijcai/YuQLG17} also uses the min-max strategy from GANs, generating data around the decision boundary between known and unknown samples as unknown. \cite{neal2018open} adds another encoder network to traditional GANs to map from images to a latent space. \cite{jo2018open} generates unknown samples by marginal denoising autoencoder that provided a target distribution which is $m$ away from the distribution of known samples. \cite{DBLP:conf/iclr/LeeLLS18} generates "boundary" samples in the low-density area of in-distribution acting as unknown samples, and jointly trains confident classifier and adversarial generator to make both models improve each other. Generating unknown samples for the OSR problem has achieved great performance, meanwhile, it requires more complex network architectures.
\begin{figure*}
\includegraphics[height=2in, width=6in]{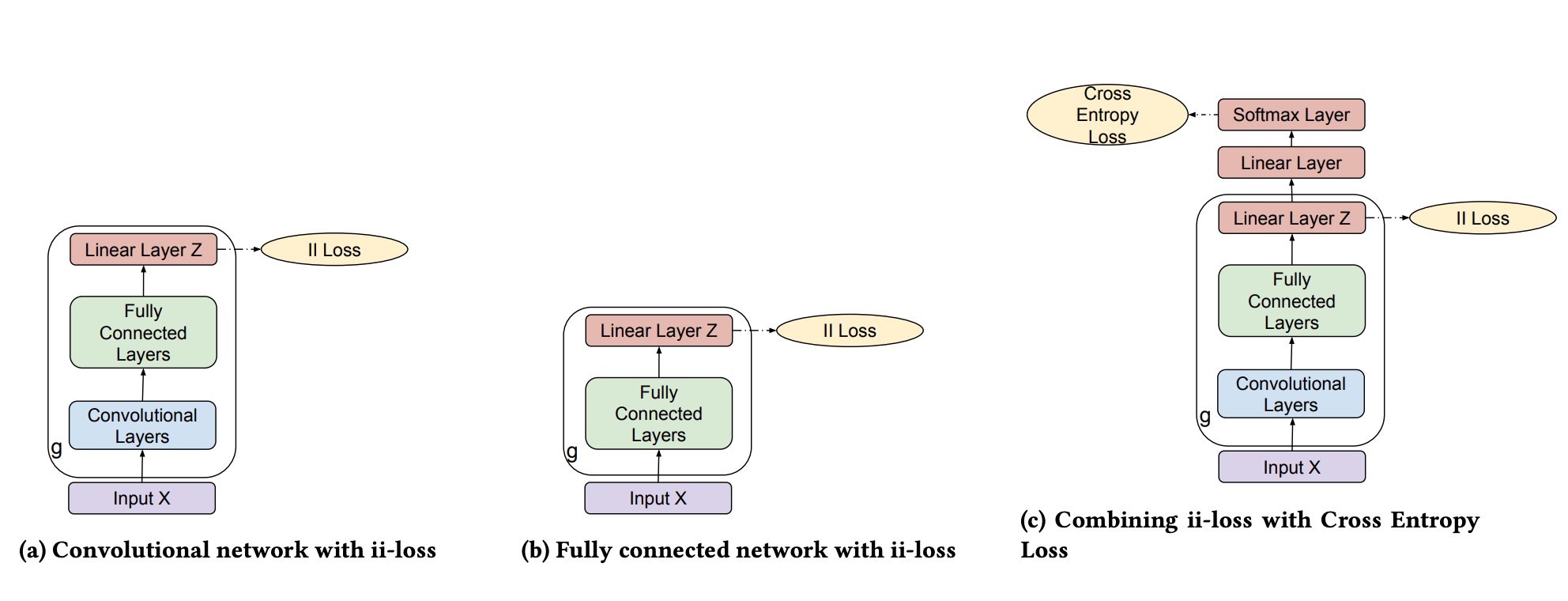}\Description{ii-loss}
\caption{Network Architecture of ii-loss (\citeauthor{hassen2018learning}, 2018)}
\label{fig:iiloss}
\end{figure*}

\subsection{No additional data}
The OSR techniques not requiring additional data in training can be divided to DNN-based (\cite{bendale2016towards} \cite{hassen2018learning} \cite{mao2018distribution} \cite{zhang2016sparse} \cite{DBLP:conf/iclr/LiangLS18}) and traditional ML-based (\cite{junior2017nearest}). 

\subsubsection{Extreme Value Signatures}
\citet{schultheiss2017finding} investigated class-specific activation patterns to leverage CNNs to novelty detection tasks. They introduced ``extreme value signature'', which specifies which dimensions of deep neural activations have the largest value. They also assumed that a semantic category can be described by its signature. Thereby, a test example will be considered as novel if it is different from the extreme-value signatures of all known categories, They applied extreme value signatures on the top of existing models, which allow to ``upgrade'' arbitrary classification networks to jointly estimate novelty and class membership.

\subsubsection{OpenMax}
\citet{bendale2016towards} proposed OpenMax which replaces the softmax layer in DNNs with an OpenMax layer, and the model estimates the probability of an input being from an unknown class. The model adopts the Extreme Value Theory (EVT) meta-recognition calibration in the penultimate layer of the networks. For each instance, the activation vector is revised to the sum of the product of its distance to the mean activation vectors (MAV) of each class. Further, it redistributes values of activation vector acting as activation for unknown class. Finally, the new redistributed activation vectors are used for computing the probabilities of both known and unknown classes.

\subsubsection{ii-loss}
\citet{hassen2018learning} proposed a distance-based loss function in DNNs in order to learn the representation for open set recognition. The idea is to maximize the distance between different classes (inter-class separation) and minimize distance of an instance from its class mean (intra-class spread). So that in the learned representation, instances from the same class are close to each other while those from different classes are further apart. More formally, let $\overrightarrow{z_i}$ be the the projection (embedding) of the input vector $\overrightarrow{x_i}$ of instance $i$. The intra class spread is measured as the average distance of instances from their class means:

\begin{equation}\label{eq:intra}
intra\_spread = \frac{1}{N}\sum_{j=1}^{K} \sum_{i=1}^{C_j}\|\overrightarrow{\mu_j}-\overrightarrow{z_i}\|_{2}^2,
\end{equation}
where $|C_j|$ is the number of training instances in class $C_j$, $N$ is the number of training instances, and $\mu_j$ is the mean of class $C_j$. Meanwhile, the inter class separation is measured as the closest two class means among all the $K$ known classes:

\begin{equation}
inter\_separation = \min_{\substack{
   1\leq m \leq K \\
   m+1\leq n \leq K
  }} \|\overrightarrow{\mu_m}-\overrightarrow{\mu_n}\|_{2}^2
\end{equation}

The proposed ii-loss minimizes the intra-class spread and maximizes inter-class separation:
\begin{equation}
    ii\_loss = intra\_spread - inter\_separation
\end{equation}

So that the distance between an instance and the closest known class mean can be used as a criterion of unknown class. i.e. if the distance above some threshold, the instance then be recognized as an unknown class. The network architecture as Figure \ref{fig:iiloss}.

\subsubsection{Distribution networks}
\citet{mao2018distribution} assumed that through a certain mapping, all the classes followed different Gaussian distributions. They proposed a distributions parameters transfer strategy to detect and model the unknown classes through estimating those of known classes. Formally, let $\boldsymbol{z}_i^k$ denotes the embedding of $\boldsymbol{x}_i^k$, they assume samples from class $k$ follow a probability distribution $p_k(\boldsymbol{z}; \boldsymbol{\Theta}_k)$ with learnable parameters $\boldsymbol{\Theta}_k$ in the latent space. For class $k$, the log-likelihood is

\begin{equation}
    \log\mathcal{L}_k(\boldsymbol{\Theta}_k, \boldsymbol{W}) = 
    \sum_{i=1}^{n_k} \log p_k(\boldsymbol{z}_i^k; \boldsymbol{\Theta}_k)
\end{equation}

The training objective is to make samples more likely to belong to their labeled class. i.e, maximize the log-likelihood of each class with respect to their samples. Hence the negative mean log-likelihood is used as a loss function in the proposed distribution networks. 

\begin{equation}
    J(\boldsymbol{W},\boldsymbol{\Theta}) = - \sum_{k=1}^{l}\frac{1}{n_k}\log\mathcal{L}_k(\boldsymbol{\Theta}_k, \boldsymbol{W})
\end{equation}

The method can not only detect novel samples but also differentiate and model unknown classes, hence discover new patterns or even new knowledge in the real world.

\subsubsection{OSNN}
Besides DNNs, \citet{junior2017nearest} introduced OSNN as an extension for the traditional machine learning technique - Nearest Neighbors(NN) classifier. It applies the Nearest Neighbors Distance Ratio (NNDR) technique as a threshold on the ratio of similarity scores. Specifically, it measures the ratio of the distances between a sample and its nearest neighbors in two different known classes. And assign the sample to one of the class if the ratio is below a certain threshold. And samples who are ambiguous between classes (ratio above a certain threshold) and those faraway from any unknown class are classified as unknown.

\subsubsection{RLCN}
\begin{figure}
\centering
\includegraphics[width=3in]{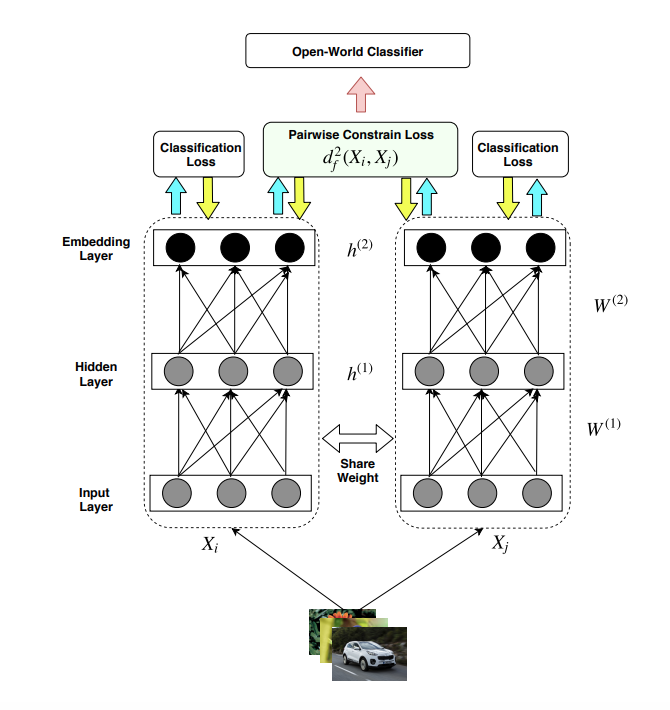}\Description{RLCN}
\caption{Overview of the RLCN Framework (\citeauthor{wang2018co}, \citeyear{wang2018co})}
\label{fig: rlcn}
\end{figure}

\citet{wang2018co} proposed a pairwise-constraint loss (PCL) function to achieve ``intra-class compactness'' and ``inter-class separation'' in order to address OSR problem. They also developed a two-channel co-representation framework to detect novel class over time. In addition to which, they added a Frobenius regularization term to avoid over-fitting. Their model also applied binary classification error(BCE) at the final output layer to form the entire loss function. Moreover, they applied temperature scaling and t distribution assumptions to find the optimal threshold, which requires fewer parameters. The two-channel co-representation framework looks like figure \ref{fig: rlcn}.

\subsubsection{SROSR}
\begin{figure*}
\includegraphics[height=2in, width=5in]{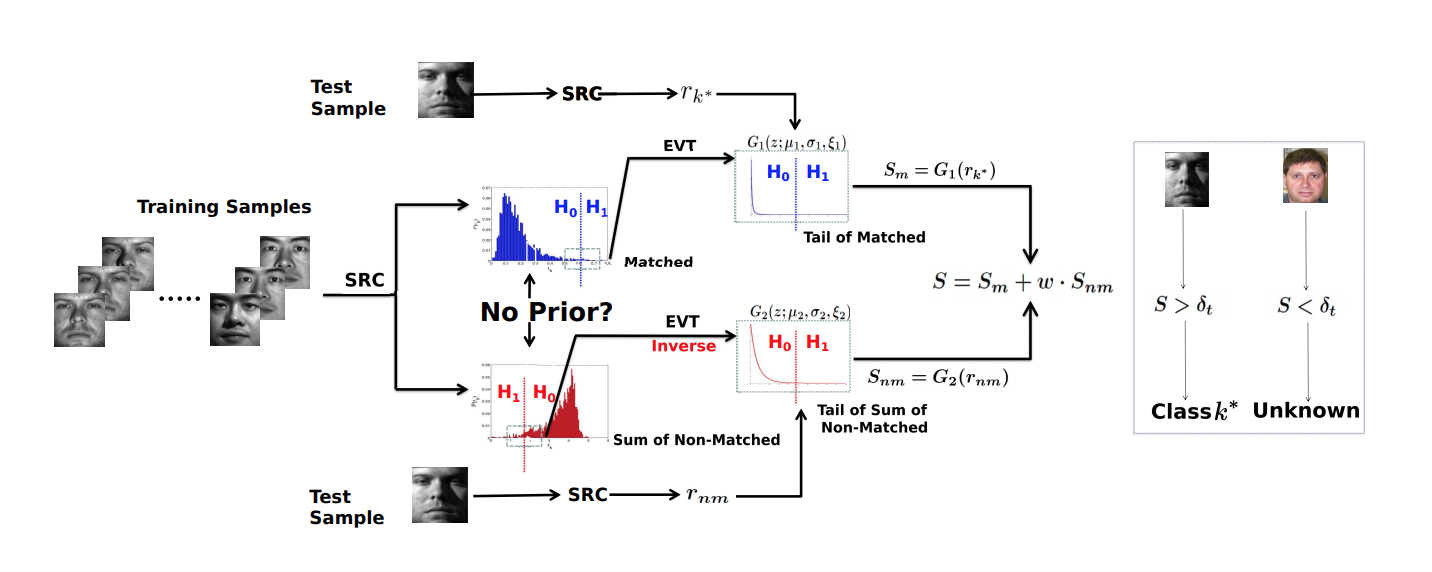}\Description{SROSR}
\caption{Overview of the SROSR framework (\citeauthor{zhang2016sparse}, \citeyear{zhang2016sparse})}
\label{fig:SROSR}
\end{figure*}

\citet{zhang2016sparse} proposed a generalized Sparse Recognition based Classification (SRC) algorithm for open set recognition in. The algorithm uses class reconstruction errors for classification. It models the tail of those two error distributions using the statistical Extreme Value Theory (EVT), then simplifies the open set recognition problem into a set of hypothesis testing problems. Figure \ref{fig:SROSR} gives an overview of the proposed SROSR algorithm.

The algorithm consists of two main stages. In the first stage, given training samples, SROSR models tail part of the matched reconstruction error distribution and the sum of non-matched reconstruction error using the EVT. In the second stage, the modeled distributions and the matched and the non-matched reconstruction errors are used to calculate the confidence scores for test samples. Then these scores are fused to obtain the final score for recognition.

\subsubsection{ODIN}
\citet{DBLP:conf/iclr/LiangLS18} proposed ODIN, an out-of-distribution detector, for solving the problem of detecting out-of-distribution images in neural networks. The proposed method does not require any change to a pre-trained neural network. The detector is based on two components: temperature scaling and input pre-processing. Specifically, ODIN set a temperature scaling parameter in original softmax output for each class like:

\begin{equation}
S_i(\mathbf{x};T) = \frac{\exp(f_i(\mathbf{x}/T))}{\sum_{j=1}^N\exp(f_j(\mathbf{x})/T)}
\end{equation}

ODIN used the maximum softmax probability softmax score, the temperature scaling can push the softmax scores of in- and out-of-distribution images further apart from each other, making the out-of-distribution images distinguishable. Meanwhile, small perturbations were added to the input during pre-processing to make in- distribution images and out-of-distribution images more separable.

\subsubsection{DOC}
\begin{figure}
\centering
\includegraphics[width=3in]{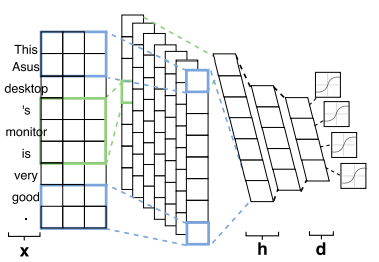}\Description{DOC}
\caption{Overview of DOC framework (\citeauthor{DBLP:conf/emnlp/ShuXL17}, \citeyear{DBLP:conf/emnlp/ShuXL17})}
\label{fig: doc}
\end{figure}

To address open classification problem, \citeauthor{DBLP:conf/emnlp/ShuXL17} proposed Deep Open Classification (DOC) method in \cite{DBLP:conf/emnlp/ShuXL17}. DOC builds a multi-class classifier with a 1-vs-rest final layer of sigmoids rather than softmax to reduce the open space risk as Figure \ref{fig: doc}. Specifically, the 1-vs-rest layer contains one sigmoid function for each class. And the objective function is the summation of all log loss of the sigmoid functions:

\begin{equation}
Loss = \sum_{i=1}^m\sum_{j=1}^n-\mathbb{I}(y_j=l_i)\log p(y_j=l_i)- \mathbb{I}(y_j\neq l_i)\log (1-p(y_j=l_i))
\end{equation}

where $\mathbb{I}$ is the indicator function and $p(y_j=l_i) = Sigmoid(d_{j, i})$ is the probability output from $i$th sigmoid function ($i$th class) on the $j$th document's $i$th dimension of $d$.

DOC also borrows the idea of outlier detection in statistics to reduce the open space risk further for rejection by tightening the decision boundaries of sigmoid functions with Gaussian fitting. It fits the predicted probability for all training data of each class, then estimates the standard deviation to find the classification thresholds for each different class.
\bigbreak

\noindent\textbf{Discussion} \\
The above papers manage to solve the OSR problems without additional datasets, and some of them adopt similar ideas as in Table \ref{tab:sim}. \cite{schultheiss2017finding}, \cite{bendale2016towards} and \cite{zhang2016sparse} utilize EVT to distinguish unknown class and known classes. \cite{hassen2018learning} and \cite{wang2018co} design different distance-based loss functions to achieve ``intra-class compactness'' and ``inter-class separation''. Some technologies are applied in different ways: \cite{wang2018co} uses temperature scaling to find the threshold of outliers, while \cite{DBLP:conf/iclr/LiangLS18} uses temperature scaling in softmax output. \cite{mao2018distribution} assumes all the classes followed different Gaussian distributions, while  \cite{DBLP:conf/emnlp/ShuXL17} tightens the decision boundaries of sigmoid functions with Gaussian fitting. In general, not using an additional dataset requires the networks generating more precise representations for known classes.  Other than DNN, \cite{junior2017nearest} introduces an extension for the Nearest Neighbors classifier.

\begin{table*}
\caption{Similarities and Differences of OSR Techniques without Additional Data}
 \label{tab:sim}
\begin{tabular}{@{}|l|l|l|l|l|l|l|l|l|l|@{}}
\toprule
Ideas  & 
\cite{schultheiss2017finding} & 
\cite{bendale2016towards} & 
\cite{hassen2018learning} & 
\cite{mao2018distribution} & 
\cite{junior2017nearest} & 
\cite{wang2018co} & 
\cite{zhang2016sparse} & 
\cite{DBLP:conf/iclr/LiangLS18} & 
\cite{DBLP:conf/emnlp/ShuXL17} \\ \midrule
DNN                              & x                                              & x                                          & x                                          & x                                           &                                           & x                                  & x                                       & x                                                & x                                               \\ 
EVT                              & x                                              & x                                          &                                            &                                             &                                           &                                    & x                                       &                                                  &                                                 \\ 
SRC                              &                                                &                                            &                                            &                                             &                                           &                                    & x                                       &                                                  &                                                 \\ 
Distance-based activation vector&                                                & x                                          &                                            &                                             &                                           &                                    &                                         &                                                  &                                                 \\ 
Distance-based loss function             &                                                &                                            & x                                          &                                             &                                           & x                                  &                                         &                                                  &                                                 \\ 
Gaussian distribution            &                                                &                                            &                                            & x                                           &                                           &                                    &                                         &                                                  & x                                               \\
Temperature scaling              &                                                &                                            &                                            &                                             &                                           & x                                  &                                         & x                                                &                                                 \\
Input perturbations              &                                                &                                            &                                            &                                             &                                           &                                    &                                         & x                                                &                                                 \\ 
1-vs-rest                        &                                                &                                            &                                            &                                             &                                           &                                    &                                         &                                                  & x                                               \\ 
Nearest neighbors                &                                                &                                            &                                            &                                             & x                                         &                                    &                                         &                                                  &                                                 \\ \bottomrule
\end{tabular}
\end{table*}

\section{Learning graph representation}

\citet{DBLP:journals/debu/HamiltonYL17} provided a review of techniques in representation learning on graphs, which including matrix factorization-based methods, random-walk based algorithms ad graph network.

The paper introduced methods for vertex embedding and subgraph embedding. The vertex embedding can be viewed as encoding nodes into a latent space from an encoder-decoder perspective. The goal of subgraph embedding is to encode a set of nodes and edges, which is a continuous vector representation.

\subsection{Vertex embedding}

Vertex embedding can be organized as an encoder-decoder framework. An encoder maps each node to a low-dimensional vector or embedding. And decoder decodes structural information about the graph from the learned embeddings. Adopting the encoder-decoder perspective, there are four methodological components for the various node embedding methods \cite{DBLP:journals/debu/HamiltonYL17}:

\begin{itemize}
    \item A pairwise similarity function, which measures the similarity between nodes
    \item An encoder function, which generates the node embeddings
    \item A decoder function, which reconstructs pairwise similarity values from the generated embeddings
    \item A loss function, which determines how the quality of the pairwise reconstructions is evaluated to train the model
\end{itemize}

The majority of node embedding algorithm reply on shallow embedding, whose encoder function just maps nodes to vector embedding. However, these shallow embedding vectors have some drawbacks.

\begin{itemize}
    \item No parameters are shared between nodes in the encoder, which makes it computationally inefficient.
    \item Shallow embedding fails to leverage node attributes during encoding.
    \item Shallow embedding can only generate embeddings for nodes that were present during the training phase, cannot generate embeddings for previously unseen nodes without additional rounds of optimization.
\end{itemize}

Recently, several deep neural network-based approaches have been proposed to address the above issues. They used autoencoders to compress information about a node's local neighborhood.


\subsubsection{GCN}
In the work of Graph Convolutional Networks (GCN) \cite{DBLP:conf/iclr/KipfW17}, \citeauthor{DBLP:conf/iclr/KipfW17} presented encoded the graph structure directly using a neural network model and trained on a supervised target. The adjacency matrix of the graph will then allow the model to distribute gradient information from the supervised loss and will enable it to learn representations of nodes both with and without labels.

The paper first introduces a simple and well-behaved layer-wise propagation rule for neural network models which operate directly on graphs as:

\begin{equation}
    H^{(l+1)} = \sigma \left( \Tilde{D}^{-\frac{1}{2}}\Tilde{A}\Tilde{D}^{-\frac{1}{2}}H^{(l)}W^{(l)}\right)
\end{equation}
Where $A$ is the adjacency matrix of the undirected graph, $I_N$ is identity matrix. $\Tilde{A}$ is the adjacency matrix with added self-connections with added self-connections. $\Tilde{D}_{ii}=\sum_j \Tilde{A}_{ij}$ and $W^{(l)}$ is a layer-specific trainable weight matrix. $\sigma(\cdot)$ denotes an activation function and $H^{l} \in \mathbf{R}^{N\times D}$ is the matrix of activation functions in the $l^{th}$ layer. Considering a two-layer GCN as semi-supervised node classificaiton example. The pre-processing step calculates $\hat{A} = \Tilde{D}^{-\frac{1}{2}} \Tilde{A}\Tilde{D}^{-\frac{1}{2}}$, then the forward model takes the simple form:
\begin{equation}
    Z = f(X, A) = \text{softmax}(\hat{A} \text{ ReLU}(\hat{A}XW^{(0)}) W^{(1)})
\end{equation}
Here, $W^{(0)} \in \mathbf{R}^{C\times H}$ is an input-to-hidden weight matrix for a hidden layer with $H$ feature maps. $W^{(1)} \in \mathbf{R}^{H\times F}$ is a hidden-to-output weight matrix. For semi-supervised multi-class classification, the cross entropy error is evaluated over all labeled examples:
\begin{equation}
    \mathcal{L} = - \sum_{l\in \mathcal{Y}_L} \sum_{f=1}^{F} Y_{lf} \ln Z_{lf}
\end{equation}
where $\mathcal{Y}_L$ is the set of node indices that have labels.

\begin{figure*}
\centering
\includegraphics[width=6in]{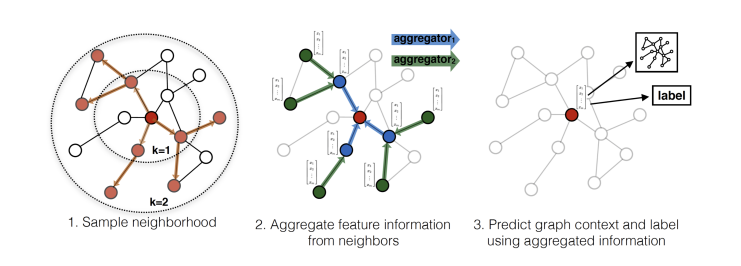}\Description{graphSAGE}
\caption{Visual illustration of the GraphSAGE sample and aggregate approach (\citeauthor{hamilton2017inductive}, \citeyear{hamilton2017inductive})}
\label{fig: graphSAGE}
\end{figure*}

\subsubsection{GraphSAGE}
Hamilton et al. presented GraphSAGE in \cite{hamilton2017inductive}, a general inductive framework that leverages node feature information to efficiently generate node embeddings for previously unseen data. GraphSAGE can learn a function that generates embeddings by sampling and aggregating features from a node's local neighborhood as Figure \ref{fig: graphSAGE}. Instead of training individual embeddings for each node, a set of aggregator functions are learned to aggregate feature information from a node's local neighborhood from a different number of hops away from a given node, for example, for aggregator function $k$ we have:

\begin{equation}
    h_{\mathcal{N}(v)}^k \leftarrow \text{AGGREGATE}_k(\{h_u^{k-1}, \forall u \in \mathcal{N}(v)\}),
\end{equation}
where $h$ is representation vector, $v$ is input node, $\mathcal{N}$ is neighborhood function. GraphSAGE then concatenates the node's current representation, $h_v^{k-1}$, with the aggregated neighborhood vector. $h_{\mathcal{N}(v)}^{k-1}$, and this concatenated vector is fed through a fully connected layer with nonlinear activation function $\sigma$ like:
\begin{equation}
    h_v^k \leftarrow \sigma \left( \mathbf{W}^k \cdot \text{CONCAT}(h_v^{k-1}, h_{\mathcal{N}(v)}^k) \right)
\end{equation}

The learned aggregation functions are then applied to the entire unseen nodes to generate embeddings during the test phase.

\subsubsection{LINE}

\begin{figure}
\centering
\includegraphics[width=3in]{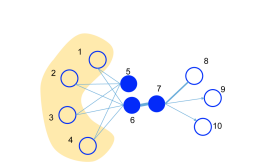}\Description{Line}
\caption{A toy example of information network (\citeauthor{tang2015line}, \citeyear{tang2015line})}
\label{fig: line}
\end{figure}

Tang et al. proposed a method for Large-scale Information Network Embedding: Line in \cite{tang2015line}, which is suitable for undirected, directed and/or weighted networks. The model optimizes an objective which preserves both the local and global network structures. The paper explores both first-order and second-order proximity between the vertices. Most existing graph embedding are designed to preserve first-order proximity, which is presented by observed links like vertex 6 and 7 in Figure \ref{fig: line}, the objective function to preserve first-order proximity looks like:

\begin{equation}
    O_1 = - \sum_{(i, j) \in E} w_{ij} \log p_1(v_i, v_j),
\end{equation}
where $p_1$ the joint probability between two vetices and is only valid for undirected edge $(i, j)$.
Besides, LINE explores the second-order proximity between the vertices, which is not determined through the observed tie strength but through the shared neighborhood structures of the vertices, such as vertex 5 and 6 should also be placed close as they share similar neighbors. In second-order proximity, each vertex is treated as a specific "context" and vertices with similar distributions over the "contexts" are assumed to be similar. To preserve the second-order proximity, LINE minimize the following objective function:

\begin{equation}
    O_2 = - \sum_{(i, j) \in E} w_{ij} \log p_2(v_j | v_i),
\end{equation}
where $p_2$ is defined as the probability of "context" $v_j$ generated by vertex $v_i$ for each directed edge $(i,j)$.

The functions preserved first-order proximity and second-order proximity are trained separately and the embeddings trained by two methods are concatenated for each vertex.

\subsubsection{JK-Net}
In order to overcome the limitations of neighborhood aggregation schemes, Xu et. al proposed Jumping Knowledge (JK) Networks strategy in \cite{DBLP:conf/icml/XuLTSKJ18} that flexibly leverages different neighborhood ranges to enable better structure-aware representation for each node. This architecture selectively combines different aggregations at the last layer, i.e., the representations “jump” to the last layer.

\begin{figure}
\centering
\includegraphics[width=3in]{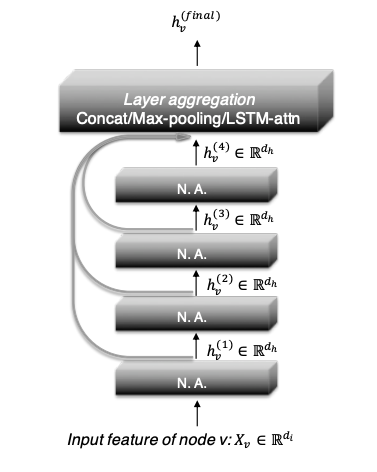}\Description{JK-Net}
\caption{Illustration of a 4-layer JK-Net. N.A. stands for neighborhood aggregation (\citeauthor{DBLP:conf/icml/XuLTSKJ18}, \citeyear{DBLP:conf/icml/XuLTSKJ18})}
\label{fig: JK-Net}
\end{figure}

The main idea of JK-Net is illustrated as Figure \ref{fig: JK-Net}: as in common neighborhood aggregation networks, each layer increases the size of the influence distribution by aggregating neighborhoods
from the previous layer. At the last layer, for each node, JK-Net selects from all of those intermediate representations (which “jump” to the last layer), potentially combining a few. If this is done independently for each node, then the model can adapt the effective neighborhood size for each node as needed, resulting in exactly the desired adaptivity. As a more general framework, JK-Net admits general layer-wise aggregation models and enable better structure-aware representations on graphs with complex structures.

\begin{figure*}
\centering
\includegraphics[width=6in]{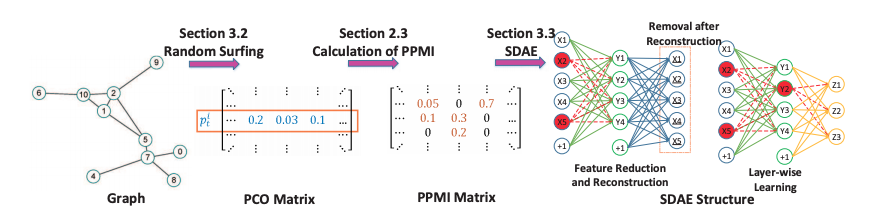}\Description{DNGR}
\caption{Main components of DNGR: random surfing, PPMI and SDAE (\citeauthor{cao2016deep}, \citeyear{cao2016deep})}
\label{fig: DNGR}
\end{figure*}

\subsubsection{DNGR}
In \cite{cao2016deep}, Cao et al. adopted a random surfing model to capture graph structural information directly instead of using a sampling-based method. As illustrated in Figure \ref{fig: DNGR}, the proposed DNGR model contains three major components: random surfing, calculation of PPMI matrix and feature reduction by SDAE.
The random surfing model is motivated by the PageRank model and is used to capture graph structural information and generate a probabilistic co-occurrence matrix. 

Random surfing first randomly orders the vertices in a graph and assume there is a transition matrix that captures the transition probabilities between different vertices. The proposed random surfing model allows contextual information to be weighted differently based on their distance to target. The generated co-occurrence matrix then used to calculate PPMI matrix (an improvement for pointwise mutual information PMI, details in \cite{levy2014neural}). Next, as high dimensional input data often contain redundant information and noise stacked denoising autoencoder (SDAE) is used to enhance the robustness of DNN, denoising autoencoder partially corrupt the input data before taking the training step. Specifically, it corrupts each input sample x randomly by assigning some of the entries in the vector to 0 with a certain probability.

\subsection{Subgraph embedding}
The goal of embedding subgraphs is to encode a set of nodes and edges into a low-dimensional vector embedding. Representation learning on subgraphs is closely related to the design of graph kernels, which define a distance measure between subgraphs. According to \cite{DBLP:journals/debu/HamiltonYL17}, some subgraph embedding techniques use the convolutional neighborhood aggregation idea to generate embeddings for nodes then use additional modules to aggregate sets of node embeddings to subgraph, such as sum-based approaches, graph-coarsening approaches. Besides, there is some related work on "graph neural networks" (GNN). Instead of aggregating information from neighbors, GNN uses backpropagation "passing information" between nodes.
\section{Malware Classification}

\subsubsection{FCG} 
\begin{figure*}
\centering
\includegraphics[width=6in]{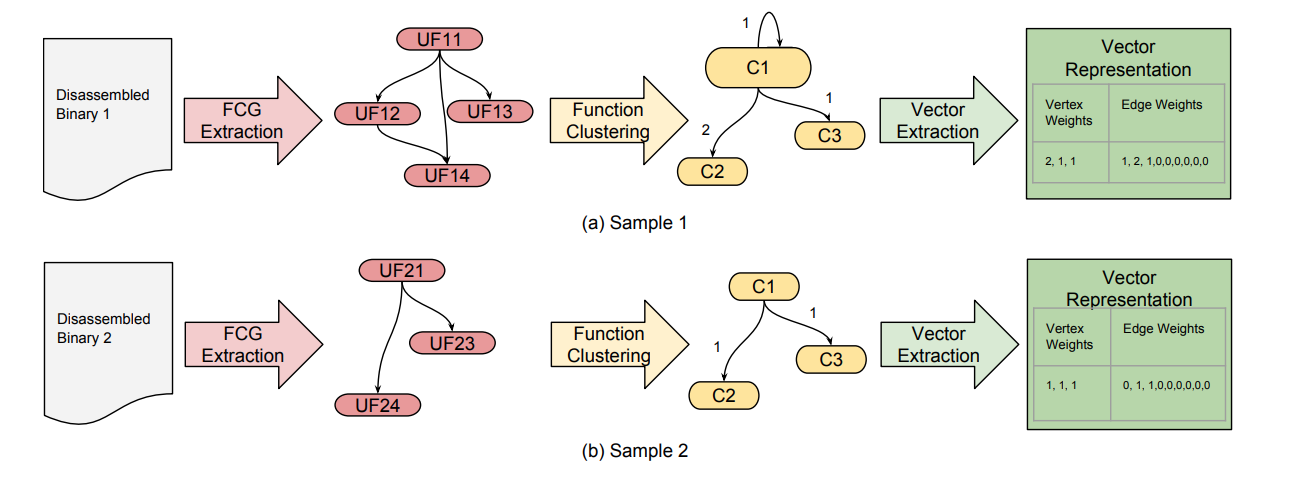}\Description{fcg}
\caption{FCG Example (\citeauthor{hassen2018learningcase}, 2018)}
\label{fig: fcg}
\end{figure*}

\citet{hassen2017scalable} proposed a linear time function call graph representation (FCG) vector representation. It starts with an FCG extraction module, which is a directed graph representation of code where the vertices of the graph correspond to functions and the directed edges represent the caller-callee relation between the function nodes. This module takes disassembled malware binaries and extract FCG representations. Thus they presented the caller-callee relation between functions as directed, unweighted edges. The next module is the function clustering. The algorithm used minhash to approximate Jaccard Index, to cluster functions of the given graph. The following module is vector extraction. The algorithm extracted vector representation from an FCG labeled using the cluster-ids. The representation consists of two parts, vertex weight, and edge weight. The vertex weight specifies the number of vertices in each cluster for that FCG and the edge weight describes the number of times an edge is found from one cluster to another cluster. The example workflow looks like figure \ref{fig: fcg}.

\subsubsection{COW, COW\rule{0.2cm}{0.15mm}PC}
Based on the work in \cite{hassen2017scalable}, \citet{hassen2018learningcase} further introduced two new features: $P_{max}$, which is the maximum predicted class probability for one instance:

\begin{equation}
P_{max} = \max_{c\in C^k}Pr(y_i=c|\Vec{x_i})
\end{equation}

And the entropy for probability distribution over classes:

\begin{equation}
entropy(p) = - \sum_j^{|C^k|}p_jlogp_j
\end{equation}

The paper also introduced two algorithms: Classification in an Open World (COW) and COW\rule{0.2cm}{0.15mm}PC. Both consist of two classifiers: outlier detector $M_{outlier}$ and multi-class classifier. The difference is in COW, the outlier detector was trained by all the classes. And during testing, test data will go through outlier detector first, if it is recognized as not outlier, then it will be sent in a multi-class classifier. While COW\rule{0.2cm}{0.15mm}PC has a class-specific outlier detector, i.e. each class has its own outlier detector. The test data will come through a multi-class classifier first, then will be sent into the corresponding outlier detector afterward.

\subsubsection{Random projections} 
Malware classifiers often use sparse binary features, and there can be hundreds of millions of potential features. In \cite{dahl2013large}, Dahl et al. used random projections to reduce the dimensionality of the original input space of neural networks. They first extracted three types of features including null-terminated patterns observed in the process’ memory, tri-grams of system API calls, and distinct combinations of a single system API call and one input parameter, next performed feature selection, ended with generating over 179 thousand sparse binary features. To make the problem more manageable, they projected each input vector into a much lower dimensional space using a sparse project matrix with entries sample iid from a distribution
over ${0,1,-1}$. Entries of 1 and -1 are equiprobable and $P(R_{ij}=0) = 1 - \frac{1}{\sqrt{d}}$, where $d$ is the original input dimensionality. The lower-dimensional data then serves as input to the neural network.

\section{Conclusions}
We provide a brief introduction of several deep neural network structures, and an overview of existing OSR, a discussion on learning graph representation and malware classification in this survey. It can be seen that those topics are advancing and profiting from each other in different areas. Also, despite the achieved great success, there are still serious challenges and great potential for them.
\bibliographystyle{ACM-Reference-Format}
\bibliography{sample-sigconf}

\end{document}